\documentclass[letterpaper, 10 pt, conference]{ieeeconf}  

\IEEEoverridecommandlockouts                              

\usepackage{graphicx} 
\usepackage{amsmath} 
\usepackage{amssymb}  
\usepackage{xurl}
\usepackage[hidelinks]{hyperref}

\title{\LARGE \bf
A Tendon-Driven Robotic Jellyfish with Constrained Soft Actuation and Depth Control via Reinforcement Learning
}

\author{Jiarui Peng$^{1,2}$, Yutong Wu$^{1,2}$, Zelong Wang$^{3}$,
        Ping Deng$^{3}$, Xiaotian Zhang$^{3}$,\\
        Xian Chen$^{3}$, Kecheng Qin$^{1,2,*}$, and Zhongyu Li$^{1,2,*}$%
\thanks{$^{1}$Hong Kong Embodied AI Lab, Hong Kong SAR, China}
\thanks{$^{2}$The Chinese University of Hong Kong, Hong Kong SAR, China}
\thanks{$^{3}$The Hong Kong University of Science and Technology, Hong Kong SAR, China}
\thanks{$^{*}$Corresponding authors: \{zhongyuli, kechengqin\}@cuhk.edu.hk}%
}

\begin{document}

\maketitle
\thispagestyle{empty}
\pagestyle{empty}

\begin{abstract}

Jellyfish-inspired robots offer a compliant and efficient approach to underwater locomotion, but achieving large deformation together with repeatable actuation and closed-loop control remains challenging. In this work, we present a tendon-driven robotic jellyfish with constrained soft actuation. Each actuator combines a flexible substrate with discrete constraints, enabling bending up to \(150^\circ\) with an approximately linear tendon displacement–bending relationship. Eight actuators driven by four servos allow the robot to perform stable swimming, attitude adjustment, and self-righting. Based on the linear actuation, a reinforcement-learning controller is further developed, enabling closed-loop depth regulation in both simulation and physical experiments. These results show that mechanical constraints can improve the controllability of soft actuation while preserving compliant jellyfish-like motion, providing a route toward manoeuvrable and autonomous jellyfish robots.

\end{abstract}

\section{Introduction}
\label{chap:intro}

Jellyfish-inspired robots provide a compliant platform for underwater observation and environmental monitoring, including applications in sensitive environments such as coral reefs and aquatic ecosystems \cite{jelly-z-environmental-monitoring}. Their propulsion relies on cyclic deformation of a compliant bell \cite{robojelly-sma}. However, reproducing this behaviour remains challenging because underwater soft actuators must combine large deformation, compliance, and repeatability \cite{soft-underwater-actuation}. Designing a soft body whose deformation remains predictable under repeated actuation is therefore a fundamental mechanical challenge.

Control is further complicated by the coupled non-linear dynamics of the deformable body and surrounding fluid. Although high-fidelity computational fluid dynamics can resolve such interactions relatively reproducibly, the computational cost makes direct real-time application impractical. Simpler controllers instead rely on reduced or experimentally identified dynamics, often requiring physical calibration and remaining sensitive to model mismatch \cite{differentiable-underwater-control} or environmental changes. Therefore, the engineering problem therein is not only to generate jellyfish-like deformation, but to make the deformation sufficiently predictable and reproducible for swimming and closed-loop control.

\begin{figure}[htbp]
    \centering
    \includegraphics[width=\columnwidth]{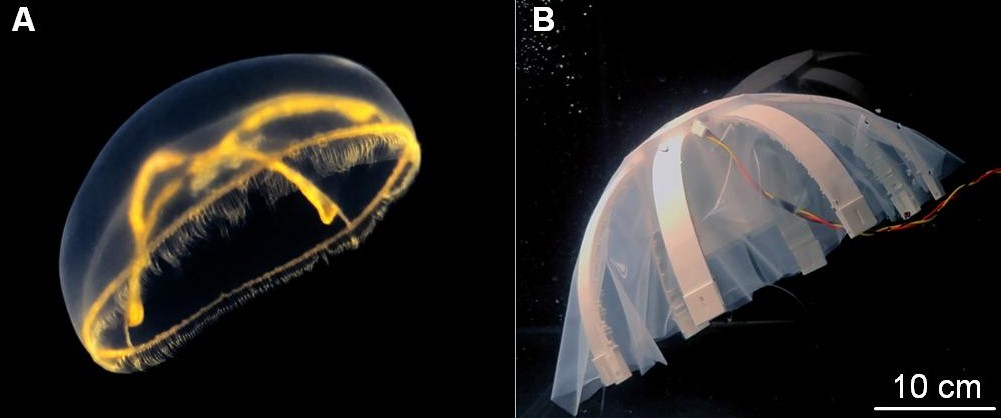}
    \caption{Biological inspiration and robotic implementation. (A) Cross jelly (\textit{Mitrocoma cellularia}), showing the compliant bell, radial canals, and densely distributed marginal tentacles. (B) Proposed tendon-driven jellyfish-inspired robot, with eight radial members supporting the compliant silicone bell. Scale bar: 10~cm.}
    \label{fig:jellyfish}
\end{figure}

The proposed robot, shown in \autoref{fig:jellyfish}, addresses both difficulties by combining constrained tendon-driven bending with a prescribed contraction-recovery stroke and a cycle-synchronous reinforcement-learning policy. Mechanical constraints regularise the deformation, while the prescribed stroke reduces the control problem to selecting a common actuation amplitude once per swimming cycle. The policy is trained entirely in simulation and transferred directly to the physical robot without additional policy training, enabling zero-shot sim-to-real closed-loop depth regulation. Attitude manoeuvring and recovery from inversion are evaluated separately under prescribed open-loop actuation.

The \textit{principal contributions} are: (1) We introduce a new constrained soft bending actuator that combines large deformation with predictable tendon-driven kinematics. A flexible thermoplastic polyurethane (TPU) substrate with discrete polylactic acid (PLA) limiting structures mechanically guides deformation, enabling approximately \(150^\circ\) bending while maintaining an approximately linear tendon-displacement–bending relationship. The proposed design ensures predictable bending angles even during large deformations. (2) We develop a novel jellyfish-inspired robotic platform using eight actuators driven by four servos for coordinated propulsion and manoeuvring. Experiments demonstrate repeatable upward swimming, asymmetric roll, pitch, and yaw manoeuvring, and recovery from inversion under prescribed pulsation. (3) We introduce a cycle-synchronous proximal policy optimisation (PPO) controller for autonomous depth regulation and the zero-shot sim-to-real demonstration of deep reinforcement learning (RL) for closed-loop depth control of a jellyfish-inspired robot. Trained in a simplified simulation, the policy selects a common actuation amplitude from depth feedback at each stroke cycle and transfers directly to the physical robot without real-world training or fine-tuning.
\section{Related Work}
\label{chap:related_work}

The related literature is reviewed from three perspectives: soft actuation, jellyfish robot systems, and modelling and~control.

\subsection{Soft Actuation for Jellyfish}
\label{sec:related-actuation}

Soft jellyfish actuators generate bell deformation through several physical mechanisms. Robojelly uses shape-memory-alloy composites to reproduce medusan morphology and kinematics \cite{robojelly-sma}, while Ren et al. developed magnetic composite lappets driven by an external oscillating magnetic field \cite{magnetic-jellyfish}. SoJel employs predominantly additively manufactured soft structures and demonstrates the sensitivity of swimming performance to material and geometric configuration \cite{sojel}. These studies highlight the coupled influence of actuation, compliance, and structure on realised bell motion. In the present work, tendon-driven bending and discrete geometric constraints instead provide a directly characterised tendon displacement-bending relationship.

\subsection{Jellyfish Robot Systems}
\label{sec:related-manoeuvrability}

Existing jellyfish-inspired robots integrate propulsion and manoeuvring through different actuation architectures. Electrohydraulic platforms employ independently actuated lappet groups to combine propulsion, steering, manipulation, and shape adaptation within a single soft robotic system \cite{electrohydraulic-jellyfish}. Pneumatic bistable actuators have also been integrated into tethered and untethered jellyfish-like prototypes to generate pulsed propulsion \cite{pneumatic-bistable-jellyfish}. In contrast, mechanically driven systems have used reciprocating and eccentric cams to separately realise propulsion and three-dimensional steering \cite{asymmetric-tentacle-steering}. These studies demonstrate different strategies for integrating actuation and locomotion functions at the system level. In the present work, four servo-driven tendon channels coordinate eight constrained soft bending actuators, providing a compact unified architecture for propulsion, manoeuvring, self-righting, and autonomous depth regulation.

\subsection{Jellyfish Modelling and Control}
\label{sec:related-feedback}

Jellyfish robot motion is influenced by hydrodynamics, which presents challenges for simulation and feedback control. Numerical studies of jellyfish locomotion range from prescribed-kinematic Computational Fluid Dynamics (CFD) to fully coupled Fluid-Structure Interaction (FSI). Simulations using measured medusan kinematics have resolved vortex formation and transport \cite{medusan-flow-transport}, while immersed-boundary and three-dimensional FSI models have coupled flexible bell deformation with viscous flow \cite{jellyfish-immersed-boundary} and reproduced turning dynamics \cite{neuromechanical-wave-resonance}. CFD has also been applied to robotic jellyfish for design and propulsion optimisation \cite{jellyfish-design-optimisation}. These methods provide detailed hydrodynamic information, but at substantially greater modelling and computational cost than reduced-order approaches \cite{immersed-boundary-fsi-review}.

Feedback control includes both conventional and learning-based methods \cite{jellyfish-millirobot-rl}. Koc et al. implemented Proportional Integral Derivative (PID) control of an artificial swim bladder for depth regulation \cite{jellyfish-pid-depth-control}. Yu et al. investigated reinforcement-learning-based attitude regulation \cite{jellyfish-rl-attitude-control}, while Wang and Chen applied reinforcement learning to jellyfish-inspired robot control \cite{jellyfish-rl-control-wang-chen}. The present work instead uses a computationally lightweight, stateless fluid approximation for repeated policy optimisation and evaluates the resulting cycle-level depth controller through physical~transfer.
\section{Robot Design and Actuator Characterisation}
\label{chap:robot_design}

This section develops the constrained bending actuators and their fabrication and integration into the robotic jellyfish. Subsequently, the characterisation of the actuator's deformation and mechanical properties provides a basis for the design of the actuation mechanism and the control system.

\subsection{Robot and Actuator Design}
\label{sec:mechanical-architecture}

\begin{figure}[htbp]
    \centering
    \includegraphics[width=\columnwidth]{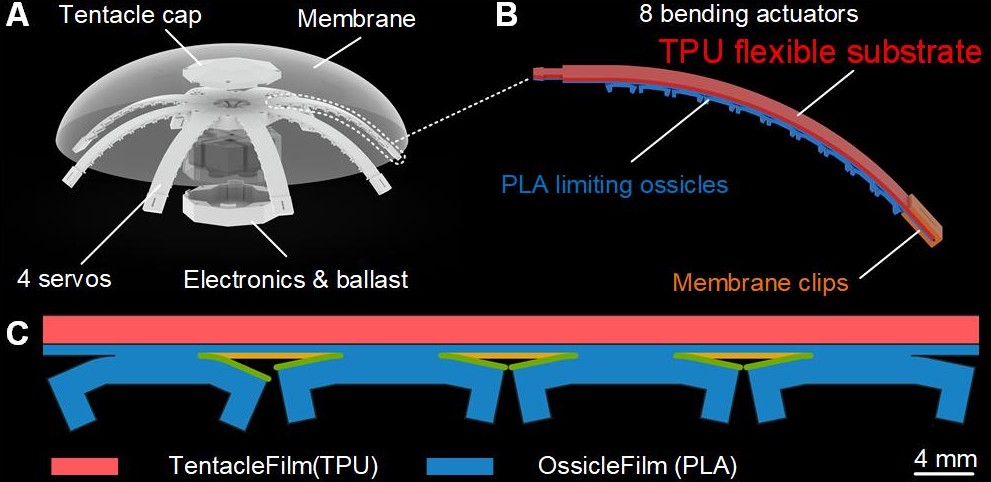}
    \caption{Mechanical architecture and actuator construction. (A) Robot 3D design model showing the bell, servomotors, radial members, electronics, and ballast. (B) Laminated radial member with a TPU 90A substrate \cite{bambu-tpu-90a}, a PLA Basic ossicle strip \cite{bambu-pla-basic}, and a distal membrane clip. (C) Flexural links and contact geometry; scale bar: 4~mm. The laminate combines localised bending with elastic recovery and geometric rotation limits.}
    \label{fig:jellyfish-structure}
\end{figure}

Inspired by jellyfish, the robot has a radially symmetric structure consisting of a central body, eight radial bending actuators, and a flexible silicone bell (\autoref{fig:jellyfish-structure}). Four servomotors, together with the electronics and ballast, are housed in the central body. Each servo drives one pair of neighbouring actuators through a shared tendon, giving four independently controlled actuation channels. The silicone bell is attached to the distal ends of the eight actuators, so their coordinated bending directly produces bell contraction and recovery. This arrangement allows a small number of servomotors to drive the entire bell while retaining independent control over different sides of the robot. Ballast is placed in the central body to adjust buoyancy and keep the centre of mass below the centre of buoyancy, providing passive attitude~restoration.

The key components of the robot are the constrained soft bending actuators. Each actuator consists of a flexible TPU substrate bonded to a series of PLA limiting structures. The TPU layer provides flexibility and elastic recovery, while the PLA structures guide the bending deformation and limit the local rotation between neighbouring segments. A tendon runs along the actuator with an offset from the TPU layer; pulling the tendon bends the actuator, while releasing it allows the TPU substrate to restore the actuator towards its original shape. The geometry of the PLA constraints is designed to produce a large but controlled bending motion, reaching approximately \(150^\circ\) in the current design.

\subsection{Actuator Characterisation}
\label{sec:programmed-deformation}

\begin{figure}[!htbp]
    \centering
    \includegraphics[width=\columnwidth]{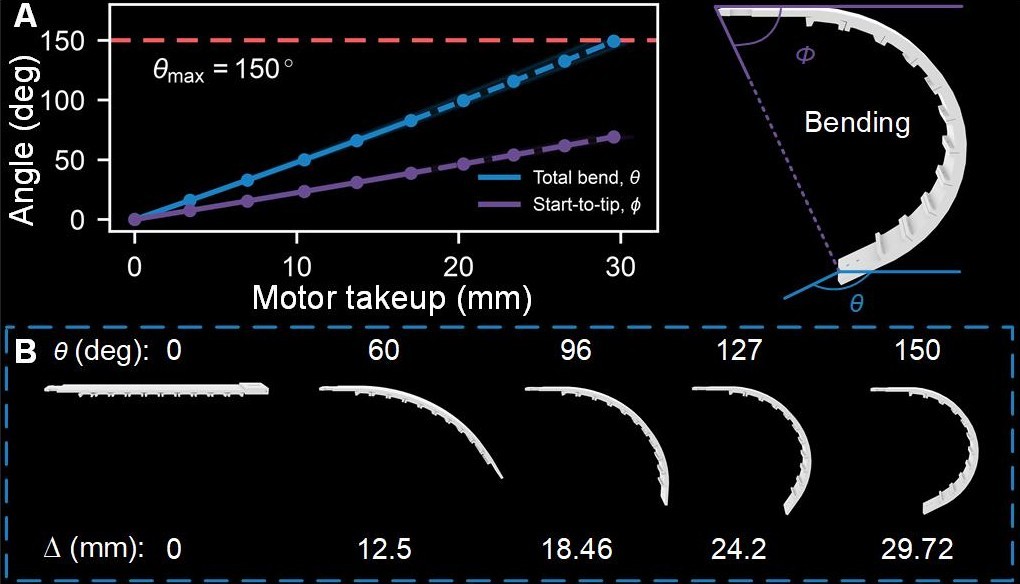}
    \caption{Bending geometry and characterisation. (A) Total bend angle \(\theta\) and start-to-tip angle \(\phi\) versus motor take-up, with geometric definitions. Central curves follow the 3D design model, while the accompanying deviations reflect physical measurements. (B) Representative configurations spanning \(0\) to \(29.72\,\mathrm{mm}\) take-up and \(0^\circ\) to approximately \(150^\circ\) total bend. The geometry and measurements support a quasi-linear displacement-bending~relationship.}
    \label{fig:skeleton-bending}
\end{figure}

We characterised the actuator to determine how tendon input is converted into bending and how much tendon force is required during actuation. \autoref{fig:skeleton-bending} shows the relationship between tendon take-up and bending angle. Both the geometric prediction and physical measurements show an approximately linear trend over the tested range. A tendon take-up of 29.72 mm produces approximately \(150^\circ\) of bending, covering the full operating range used in the robot.

Because the tendon itself stretches under load, its deformation also affects the transmission. The monofilament tendon was therefore subjected to a stretch-release procedure before installation to improve repeatability. After preconditioning, its measured compliance was approximately \(0.586~\mathrm{mm\,N^{-1}}\). The actuators also showed good durability during repeated experimental use: the same eight actuators were used for more than 100 experimental sessions without visually noticeable permanent deformation, although these observations are not intended as a formal fatigue-life evaluation.

\begin{figure}[htbp]
    \centering
    \includegraphics[width=\columnwidth]{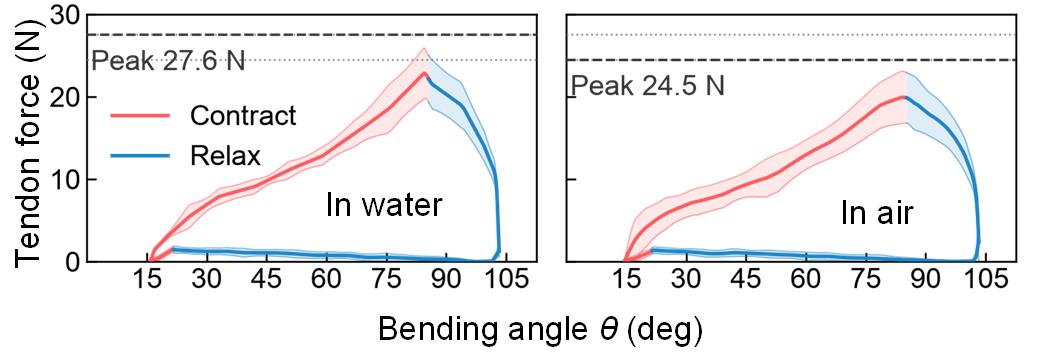}
    \caption{Tendon force during contraction and recovery in water and air. Shading indicates one standard deviation about the mean. The largest recorded forces were \(27.6\,\mathrm{N}\) in water and \(24.5\,\mathrm{N}\) in air, a \(3.1\,\mathrm{N}\) increase (approximately \(13\%\)). The comparison highlights path-dependent loading and the higher peak load during submerged actuation.}
    \label{fig:angle-force-curve}
\end{figure}

We further measured the tendon force during contraction and recovery in both air and water (\autoref{fig:angle-force-curve}). The force follows different paths during contraction and release, indicating that the actuator exhibits path-dependent loading even though its displacement–bending relationship is approximately linear. The peak tendon force was higher in water than in air, reflecting the additional load encountered during submerged actuation. This measurement characterises the load on the tendon transmission rather than the propulsive thrust generated by the robot.

\section{Open-Loop Swimming Gait}
\label{chap:openloop}

This section presents the prescribed open-loop swimming gait used by the robotic jellyfish. The four tendon-driven channels follow a synchronised contraction–recovery cycle, while their actuation amplitudes can be adjusted independently to generate different bell deformation patterns. Based on this gait, the robot achieves symmetric propulsion, asymmetric manoeuvring, and self-righting with passive mechanical~restoration.

\subsection{Swimming Gait Generation}
\label{sec:openloop-actuation}

\begin{figure}[htbp]
    \centering
    \includegraphics[width=\columnwidth]{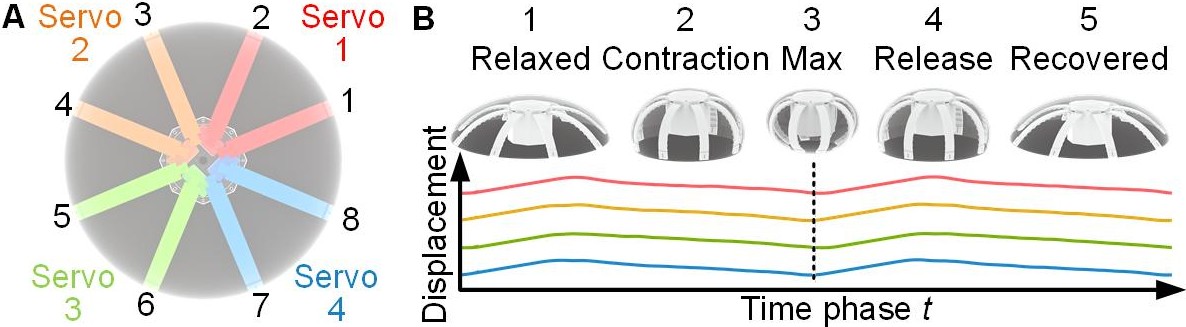}
    \caption{Prescribed open-loop actuation organisation and swimming cycle. (A) Four servomotors each actuate a neighbouring pair of radial members. (B) Representative bell configurations and schematic displacement traces during consecutive cycles. Shared stroke timing permits amplitude asymmetry while preserving phase synchronisation.}
    \label{fig:actuation-cycle}
\end{figure}

The swimming gait is generated by four independently commanded tendon channels operating with the same cycle timing. As shown in \autoref{fig:actuation-cycle}, each servomotor drives a neighbouring pair of radial actuators. All four channels share a common frequency \(f\), while the tendon-retraction amplitude \(A_i\) can be set independently for each channel.

Each cycle consists of a contraction phase followed by a recovery phase. Contraction occupies the first one-third of the cycle and recovery the remaining two-thirds, giving a contraction speed twice the recovery speed for the same stroke amplitude. When different amplitudes are commanded, the servo speeds are scaled accordingly so that all four channels remain synchronised and reach their stroke limits at the~same~time.

During contraction, tendon retraction bends the radial actuators and stores elastic energy in the structure. During recovery, the servomotor releases the tendon, allowing the actuators to return towards their relaxed shape through their own elasticity. The servomotor therefore controls the rate of reopening rather than actively driving the bell outwards. This elastic recovery is similar to the restoring role of compliant tissues in hydromedusae \cite{hydromedusa-mesoglea-elasticity}, although it is different from fluid-mediated passive energy recapture such as the stopping-vortex effect reported for \textit{Aurelia aurita} \cite{passive-energy-recapture}. The different force paths during tendon retraction and release in \autoref{chap:robot_design} further reflect this contraction–recovery behaviour.

With equal amplitudes, the bell contracts symmetrically and generates axial propulsion. Changing the amplitude of individual channels creates asymmetric bell deformation and therefore produces manoeuvring without changing the relative phase between channels. The contraction–recovery pattern follows the general organisation of medusan swimming \cite{jellyfish-hydrodynamics}. Meanwhile, the specific timing used here was selected~empirically.

For the open-loop experiments, the pulsation frequency was set to approximately \(0.4~\mathrm{Hz}\), corresponding to a \(2.5~\mathrm{s}\) cycle with about \(0.83~\mathrm{s}\) contraction and \(1.67~\mathrm{s}\) recovery. This frequency lies within the \(0.25\text{–}1.00~\mathrm{Hz}\) range reported for \textit{Aurelia aurita} \cite{jellyfish-frequency}. The maximum commanded tendon travel was limited to \(29.72~\mathrm{mm}\), corresponding to the approximately \(150^\circ\) actuator bending characterised in \autoref{fig:skeleton-bending}.

\subsection{Propulsion and Manoeuvre}
\label{sec:physical-behaviour-overview}

We illustrate propulsion, asymmetric manoeuvre, and mechanically supported righting through representative open-loop swimming sequences. \autoref{fig:swimming-experiments} summarises these capabilities, with experimental conditions and quantitative evaluation presented later.

Panels A and B show a representative synchronous ascent and the corresponding upward displacement. Panels C and D show asymmetric steering and attitude evolution; the displayed sequence illustrates part of the achievable reorientation rather than defining the maximum turning angle. Panels E and F show recovery from inversion followed by upward motion, with passive mechanical righting supported by the calibrated ballast-buoyancy arrangement during continued~pulsation.

These demonstrations establish physical capabilities before learned depth regulation, rather than providing a matched depth-holding baseline. The learned controller retains the contraction-recovery organisation, with its action mapping and amplitude-dependent cycle timing specified separately in \autoref{sec:rl_control}.
\section{Reinforcement Learning for Depth Control}
\label{sec:rl_control}

To control depth, we use a reinforcement-learning controller to adjust the swimming amplitude according to the robot’s vertical state. At the beginning of each swimming cycle, the policy selects one common amplitude for all four actuation channels, while the predefined swimming gait and passive attitude restoration remain unchanged. The policy is trained with PPO in a simplified simulation and then deployed directly on the physical robot.

\subsection{Cycle-Level Control}
\label{sec:cycle_control}

We formulate depth regulation as a Partially Observable Markov Decision Process (POMDP), since onboard sensing does not fully capture the fluid and tendon states. At decision boundary \(k\), the selected amplitude is held throughout one complete swimming cycle of realised duration \(T_k\).

Depth is positive downwards. Given target depth \(d^\ast\) and estimated depth \(\widehat d_k\), the boundary error is \(e_{d,k}=\widehat d_k-d^\ast\). The deployment-equivalent observation is
\begin{equation}
    \mathbf{o}_k =
    \begin{bmatrix}
        e_{d,k} & \widehat v_{d,k} & u_{k-1}
    \end{bmatrix}^{\mathsf T}
    \in \mathbb{R}^{3},
    \label{eq:actor_observation}
\end{equation}
where \(\widehat v_{d,k}\) is estimated downward velocity and \(u_{k-1}\) is the preceding normalised action. Positive error denotes a position deeper than the target; descent is achieved by reducing propulsion under slight negative buoyancy.

The critic receives the same observation, \(\boldsymbol{\xi}_k=\mathbf{o}_k\); neither network receives horizontal position or attitude. During training, the actor with parameters \(\theta\) samples \(u_k\sim\pi_\theta(\cdot\mid\mathbf{o}_k)\), with \(u_k\in[-1,1]\). Evaluation and deployment use deterministic actor output.

\subsection{Action Mapping and Cyclic Timing}

We map the scalar action to a hover-centred amplitude shared by all four quadrants, which contract and recover in phase. Let \(A_h\) denote the nominal simulated hovering amplitude, with admissible bounds \(A_{\min}\) and \(A_{\max}\). Defining the available spans as \(\Delta A_-=A_h-A_{\min}\) and \(\Delta A_+=A_{\max}-A_h\), the mapping is
\begin{equation}
    A_k =
    \begin{cases}
        A_h+\Delta A_-u_k, & u_k<0,\\
        A_h+\Delta A_+u_k, & u_k\geq0.
    \end{cases}
    \label{eq:amplitude_mapping}
\end{equation}
Thus \(u_k=-1,0,1\) correspond to \(A_{\min},A_h,A_{\max}\), respectively, allowing propulsion adjustments around the nominal hovering setting.

A calibrated gain \(\kappa_q\) converts amplitude into servo-coordinate excursion. Taking tendon retraction as positive, the nominal contraction target is \(q_{\mathrm r}+\kappa_q A_k\), where \(q_{\mathrm r}\) denotes the relaxed coordinate. Software position limits yield the issued target \(q_k^\ast\), with corresponding excursion \(S_k=|q_k^\ast-q_{\mathrm r}|\). This representation is independent of the servo's internal encoding convention.

The prescribed recovery and contraction rate magnitudes are \(\nu\) and \(2\nu\), respectively. Let \(\tau(S;\nu,\alpha)\) denote the travel time for excursion \(S\) under rate limit \(\nu\) and prescribed acceleration magnitude \(\alpha\). With total endpoint holds \(T_{\mathrm h}\) and controller-imposed minimum duration \(T_{\min}\), cycle timing is
\begin{equation}
    T_k =
    \max\bigl\{
        T_{\min},
        \tau(S_k;2\nu,\alpha)
        +\tau(S_k;\nu,\alpha)
        +T_{\mathrm h}
    \bigr\}.
    \label{eq:cycle_duration}
\end{equation}
Short strokes wait until the minimum duration has elapsed. Unlike the fixed-frequency open-loop trials in \autoref{chap:openloop}, cycle duration can therefore vary with amplitude, while timing rules and inter-quadrant phase remain prescribed.

\subsection{Reward and PPO Training}
\label{sec:ppo_training}

We train the policy using PPO \cite{ppo} to reduce depth error, suppress residual vertical motion, and discourage abrupt changes between successive actions. The cycle reward is
\begin{equation}
    r_k =
    -w_d\overline{e_d^{\,2}}
    -w_v\overline{v_d^{\,2}}
    -w_u(u_k-u_{k-1})^2,
    \label{eq:hover_reward}
\end{equation}
where the positive weights \(w_d,w_v,w_u\) set the relative penalties. The overbar denotes the mean over all physics steps within the completed cycle, using simulator ground-truth error \(e_d=d-d^\ast\) and downward velocity \(v_d\). Reward evaluation therefore uses the simulated trajectory, while both networks receive only the estimated boundary observation.

The critic \(V_\phi(\boldsymbol{\xi}_k)\), with parameters \(\phi\), estimates discounted return. Because complete strokes form temporally extended actions \cite{temporal-abstraction}, discounting accounts for elapsed physical time through \(\gamma_k=\exp(-\beta_\gamma T_k)\). The Generalised Advantage Estimation (GAE) trace factor is similarly set to \(\lambda_k=\exp(-\beta_\lambda T_k)\) \cite{gae}, where \(\beta_\gamma\) and \(\beta_\lambda\) are fixed positive decay rates with units of inverse time. Duration scaling of \(\lambda_k\) is an implementation choice. Time-limit truncations retain value bootstrapping, whereas true terminal transitions do not \cite{time-limits-rl}.

\subsection{Simulation and Deployment}
\label{sec:training_distribution}
\label{sec:low_order_sim}

We train in a control-oriented simulation and preserve its observation and action interfaces for physical deployment. The simulation environment \cite{mujoco} represents six-degree-of-freedom body motion, segmented radial actuators, and a discretised deformable bell. Native stateless fluid forces \cite{mujoco-fluid-forces} approximate water-robot interaction without resolving wake history. The model follows the physical trim strategy, with slight negative buoyancy and the centre of buoyancy above the centre of mass to provide passive attitude restoration.

Initial depth errors, vertical velocities, and small attitude perturbations are randomised around the target condition. Mass, buoyancy, drag, and structural parameters remain fixed. Rollouts provide completed-cycle rewards, realised durations, and termination information for PPO updates. Learning therefore adjusts the strength of an established swimming primitive rather than discovering the gait or coordinating individual quadrants.

\begin{figure}[htbp]
    \centering
    \includegraphics[width=\columnwidth]{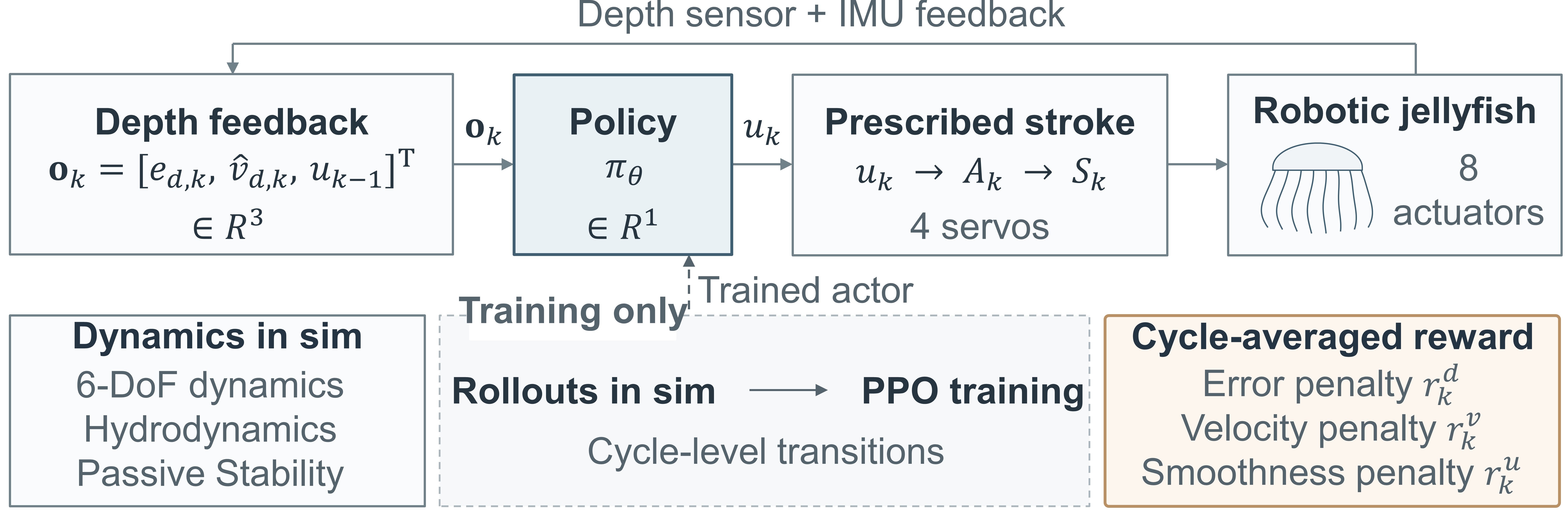}
    \caption{Cycle-synchronous RL-based depth control. Depth and inertial feedback provide the observation \(\mathbf{o}_k\); the actor selects \(u_k\), which is mapped to a common amplitude \(A_k\) for four synchronous servomotors and held throughout the stroke. Both networks share the same observation during training. Simulation rollouts, the critic, reward evaluation, and PPO optimisation are training-only components; only the actor is deployed.}
    \label{fig:reinforcement-learning-pathway}
\end{figure}

As summarised in \autoref{fig:reinforcement-learning-pathway}, deployment retains the observation convention, amplitude mapping, software bounds, and low-level timing used during training. The actor is evaluated once per cycle with fixed network parameters, providing state-dependent depth correction without online policy optimisation.
\section{Experimental Evaluation}
\label{sec:experiments}

The evaluation first establishes physical swimming capabilities under prescribed actuation through ascent, asymmetric steering, and righting experiments. RL-based depth control is then assessed against fixed-amplitude actuation in simulation and after transfer to the physical robot.

\subsection{Environment Setup}
\label{sec:experimental-conditions}

We establish a common tank environment and measurement convention for the physical experiments. Tests were conducted in a glass tank with internal dimensions of \(80\times50\times70\,\mathrm{cm}\). Water depth was approximately \(60\,\mathrm{cm}\), with modest variation between sessions. Vertical motion was estimated from hydrostatic pressure, using the initial measurement as the local depth reference.

The assembled robot was ballasted after fabrication as described in \autoref{sec:mechanical-architecture}. Ballast distribution was adjusted to reduce static roll and pitch bias while maintaining the centre of mass below the centre of buoyancy. Experiments used approximately neutral or slightly negatively buoyant trim without requiring an exact net-buoyancy value. Two slack external connections carried sensor communication and servo power and communication, respectively. Neither cable provided mechanical support, and the robot remained free to translate and rotate within the tank.

\begin{figure*}[htbp]
    \centering
    \includegraphics[width=\textwidth]{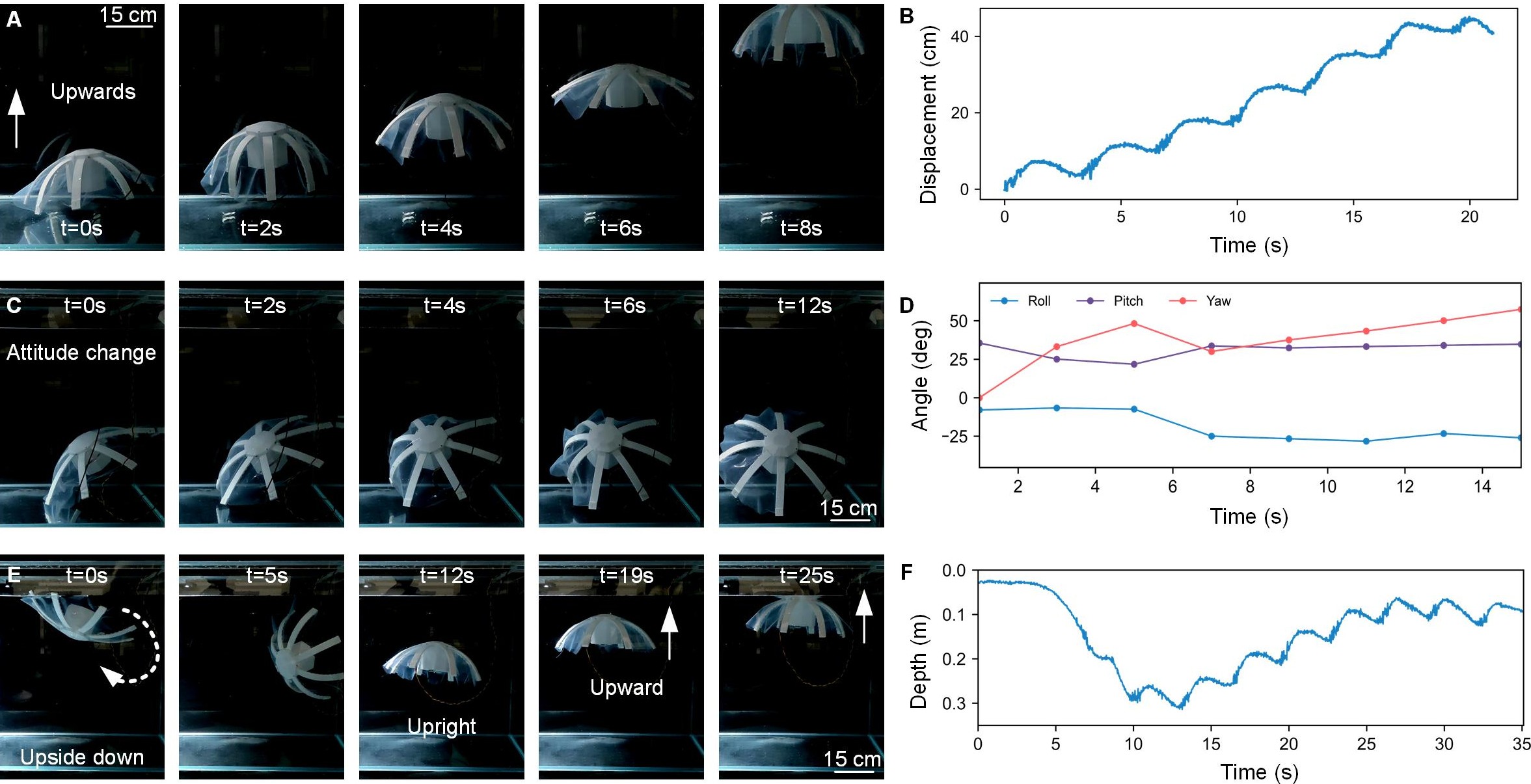}
    \caption{Physical swimming behaviours under prescribed open-loop actuation. (A) Representative synchronous ascent sequence. (B) Pressure-derived upward displacement during the representative ascent trial.
    (C) Representative asymmetric steering sequence. (D) Zero-referenced roll, pitch, and yaw evolution during the illustrated manoeuvre. (E) Recovery from inversion followed by upward motion. (F) Self-righting swimming. Righting occurred during continued pulsation without learned attitude control.}
    \label{fig:swimming-experiments}
\end{figure*}

\subsection{Open-Loop Ascent}
\label{sec:openloop-ascent}

We first test whether prescribed symmetric pulsation produces sustained propulsion without body-state feedback. All four channels were driven synchronously at \(0.4\,\mathrm{Hz}\) using the contraction-recovery cycle in \autoref{sec:openloop-actuation}. Each channel used a nominal tendon excursion of approximately \(15.7\,\mathrm{mm}\); tendon compliance reduced the realised excursion slightly to approximately \(15.4\,\mathrm{mm}\).

\autoref{fig:swimming-experiments}(A) shows the initial \(8\,\mathrm{s}\) of a representative ascent, while panel B presents the corresponding pressure-derived ascent trace extending beyond the photographed interval. The panels share the same \(t=0\) reference but span different durations. Upward displacement is expressed as \(\Delta z(t)=-[\widehat d(t)-\widehat d(0)]\), where \(\widehat d(t)\) denotes downward-positive pressure-derived depth. Repeated synchronous pulsation produced sustained upward displacement, demonstrating that the compliant bell and programmed radial deformation provide sufficient propulsion for unsupported swimming.

\subsection{Steering and Self-Righting}
\label{sec:steering-righting}

We examine directional authority through asymmetric actuation and mechanically supported recovery from inversion. In the illustrated steering manoeuvre, the stronger and weaker sides used different excursions for steering. The resulting attitude change is shown in \autoref{fig:swimming-experiments}(C, D). Euler-angle traces were zero-referenced for visualisation, so their relative evolution rather than their absolute initial values is interpreted. The experiment demonstrates open-loop steering through asymmetric bell deformation without a separate rudder or steering surface. The righting trial in \autoref{fig:swimming-experiments}(E, F) begins with the robot inverted. During continued bell pulsation, the robot returns towards an upright orientation and resumes upward swimming. 

\subsection{Depth Control in Simulation}
\label{sec:simulation-evaluation}

We quantify the benefit of learned depth feedback by comparing the PPO controller in \autoref{sec:rl_control} with fixed-amplitude actuation at \(A_k\equiv A_h\), where \(A_h\) denotes the nominal simulated hovering amplitude. Both controllers were evaluated from the same set of 16 initial conditions. Performance was assessed using the Root-Mean-Square Error (RMSE) of depth error \(e_d(t)=d(t)-d^\ast\), where \(d(t)\) and \(d^\ast\) denote simulated and target depth.

\begin{figure}[htbp]
    \centering
    \includegraphics[width=\columnwidth]{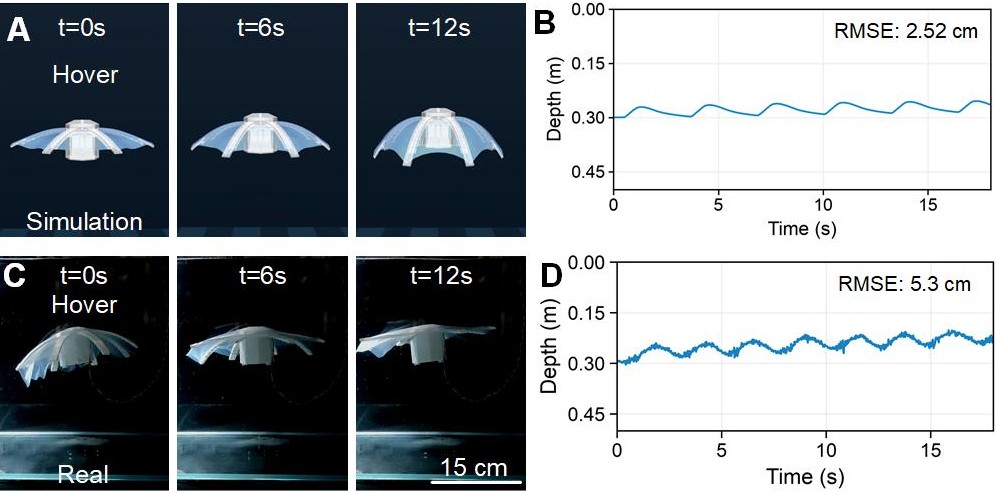}
    \caption{Representative closed-loop depth-control behaviour using the common-amplitude PPO policy.
    (A) Simulated swimming sequence.
    (B) Corresponding simulated depth.
    (C) Physical deployment of the same policy.
    (D) Corresponding pressure-derived physical depth response.
    Annotated RMSEs refer to the illustrated trajectories rather than multi-trial aggregates. These examples show closed-loop depth regulation in simulation and after transfer to the physical robot.}
    \label{fig:simulation-reality}
\end{figure}

Across the simulated test conditions, the fixed-amplitude open-loop baseline produced a depth RMSE of 8.31 cm, whereas the PPO controller reduced the aggregate RMSE to 2.83 cm, corresponding to a reduction of approximately 66\%. This result shows that state-dependent adjustment of the swimming amplitude provides substantially better depth regulation than fixed-amplitude actuation. \autoref{fig:simulation-reality}(A, B) shows one representative simulated trajectory. The RMSE annotated in \autoref{fig:simulation-reality}(B) corresponds to the illustrated episode rather than the aggregate result.

\subsection{Depth Control on the Robot}
\label{sec:physical-depth-control}

We test physical transfer by deploying the selected actor with the same one-dimensional action interface. Pressure and inertial measurements were processed using the deployment interface in \autoref{fig:reinforcement-learning-pathway}, and the actor selected one common contraction amplitude at each cycle boundary. Physical depth and its target were expressed relative to the same initial pressure reference.

The trial in \autoref{fig:simulation-reality}(C, D) was not tuned to an exact neutral-buoyancy equilibrium; a small residual ballast-buoyancy imbalance remained after trim. Nevertheless, the transferred policy varied the common stroke amplitude in response to the measured vertical state and maintained depth regulation. The RMSE annotated in panel D characterises this representative trial rather than a population statistic.

Discrepancies from simulation may reflect the stateless fluid approximation in \autoref{sec:low_order_sim}, fabrication and buoyancy mismatch, compliant tendon transmission, and confined-tank hydrodynamics. Their individual contributions were not isolated, and their effects remain included in the measured physical response.

Qualitative disturbance trials additionally showed repeated recovery after manual displacement or large attitude perturbations. Because the applied forces and moments were not instrumented, these observations are not presented as quantitative disturbance-rejection results. The physical evidence supports transfer of state-dependent vertical correction from simulation to the complete compliant robot, while passive mechanical trim remains the principal attitude-restoring~mechanism.
\section{Discussion}
\label{sec:discussion}

This section interprets the design and control choices underlying the demonstrated behaviours, then identifies modelling limitations and practical priorities for replication.

\subsection{Constrained Compliance and Passive Stabilisation}
\label{sec:discussion-morphology}

The principal design implication is that compliant deformation can be organised through local structural constraints. Bending is concentrated predominantly in the thin PLA links between ossicles, while the TPU substrate accommodates extension and supplies elastic recovery. The selected bend range reflects body dimensions rather than an intrinsic architectural limit; the total bending angle, actuator length, and curvature can therefore be customised for different designs. Non-uniform ossicle spacing, dimensions, and contact angles could extend this approach towards geometrically programmed curvature distributions through mechanics-based optimisation, subject to strain and contact constraints.

The mechanical trim restores attitude, prescribed strokes generate swimming, and the policy regulates depth. The actor need not observe or command individual segment deformations, although those mechanics still determine the resulting motion. The contribution is therefore a restricted learning interface built on established physical behaviours, consistent with using morphology to substantially reduce active-control~requirements~\cite{self-organisation-embodied-robotics}.

\subsection{Sim-to-Real Transfer}
\label{sec:discussion-transfer}

The transfer result demonstrates useful feedback despite model mismatch, rather than hydrodynamic equivalence. The simulation's native fluid approximation computes instantaneous loads without resolving the evolving wake \cite{mujoco-fluid-forces}, omitting vortex-mediated mechanisms known to contribute to medusan propulsion \cite{passive-energy-recapture}. Simulated trajectories therefore cannot establish the physical thrust mechanisms or propulsion efficiency.

The simulated baseline comparison and physical deployment provide distinct evidence: improved tracking within the model and useful depth correction on hardware. They do not establish equal performance across environments or consistently precise hovering. A targeted extension would retain the mechanical simulator while incorporating history-dependent force surrogates informed by CFD or physical measurements. Such models would require validation across stroke conditions; static force tables alone would not recover wake memory.

\subsection{Limitations and Future Work}
\label{sec:discussion-limitations}

Replication also depends on transmission durability and reproducible submerged trim. Tendon wear and occasional breakage motivate abrasion-resistant lines and improved guides, with compliance and overload behaviour reassessed after replacement. Preconditioning and post-immersion trim checks remain important: ballast corrects initial imbalance but does not prevent subsequent changes associated with trapped air or water ingress.

Longer trials in a larger volume, independent motion measurements, calibrated disturbances, and conventional feedback baselines using the same amplitude interface should precede broader robustness claims. Current tank confinement and uninstrumented disturbances limit conclusions about sustained station keeping and disturbance rejection.
\section{Conclusion}
\label{sec:conclusion}

This work presents a tendon-driven robotic jellyfish built around constrained soft bending actuators. The TPU–PLA actuators combine large, repeatable bending with elastic recovery and an approximately linear tendon displacement–bending relationship. Four servomotors drive eight radial actuators to generate upward swimming, asymmetric manoeuvring, and mechanically supported self-righting. Building on this repeatable swimming gait, a cycle-level PPO controller adjusts a single common stroke amplitude for closed-loop depth regulation. The policy improves depth tracking over fixed-amplitude actuation in simulation and transfers directly to the physical robot. Overall, the results show that constraining the actuator mechanics and prescribing the basic swimming gait can simplify the control problem, allowing learning to focus on depth regulation rather than low-level gait generation or individual actuator~coordination.





\section*{ACKNOWLEDGMENT}

This work was in part supported by the InnoHK initiative of the Innovation and Technology Commission of the Hong Kong Special Administrative Region Government via the Hong Kong Centre for Logistics Robotics. This work was also in part supported by the Chinese University of Hong Kong, The Hong Kong University of Science and Technology, and National Natural Science Foundation of China.

ChatGPT was used for language proofreading and assistance with debugging research code. All scientific content, methodology, analysis, results, and conclusions were developed and verified by the authors.

\addtolength{\textheight}{-7cm}   

\end{document}